%% file: main.tex
\documentclass[conference]{IEEEtran}
\usepackage{graphicx} 
\usepackage{cite}
\usepackage{amsmath,amssymb,amsfonts}
\usepackage{algorithmic}
\usepackage{graphicx}
\usepackage{textcomp}
\usepackage{xcolor}
\usepackage{multicol}
\usepackage{pgfplots}
\usepackage{multirow}
\usepackage{hyperref}
\usepackage{tabularx}
\usepackage{float}
\usepackage[table]{xcolor}
\usepackage{subcaption}
\pgfplotsset{compat=1.18}
\usepackage{comment}
\usepackage{svg}
\usepackage{tabularx}
\usepackage{balance}

\begin{document}
\title{Hardware optimization on Android for inference of AI models}

\author{\IEEEauthorblockN{Iulius Gherasim}\\
\IEEEauthorblockA{\textit{Department of Computer Architecture and Automation} \\
\textit{Complutense University of Madrid}\\
Madrid, Spain \\
0009-0009-2479-0960}
\and
\IEEEauthorblockN{Carlos García Sánchez}\\
\IEEEauthorblockA{\textit{Department of Computer Architecture and Automation} \\
\textit{Complutense University of Madrid}\\
Madrid, Spain \\
0000-0002-3470-1097}
}

\maketitle
\begin{abstract}
    The pervasive integration of Artificial Intelligence models into contemporary mobile computing is notable across numerous use cases, from virtual assistants to advanced image processing. Optimizing the mobile user experience involves minimal latency and high responsiveness from deployed AI models with challenges from execution strategies that fully leverage real time constraints to the exploitation of heterogeneous hardware architecture. In this paper, we research and propose the optimal execution configurations for AI models on an Android system, focusing on two critical tasks: object detection (YOLO family) and image classification (ResNet). These configurations evaluate various model quantization schemes and the utilization of on-device accelerators, specifically the GPU and NPU. Our core objective is to empirically determine the combination that achieves the best trade-off between minimal accuracy degradation and maximal inference speed-up.

\end{abstract}

\section{Introduction}
\input{intro}

\section{Environment}
\label{sec:env}

\subsection{Hardware aspects}
All the experimentation has been carried out in a stock Android tablet device, namely a Samsung Galaxy Tab S9. This device has been chosen as it comes with a System-of-Chip (SoC) which includes a NPU alongside the GPU and CPU, the devices that will be evaluated in this paper. The specific SoC is a Qualcomm Snapdragon SM8550-AC also referred to as Snapdragon 8 Gen 2 Mobile Platform. There are the three main computing units inside the SoC of interest for this paper, CPU, GPU and NPU, alongside other smaller dedicated units for networking or image processing. 

The SoC CPU, named \textit{Kryo}, is made up of eight cores arranged in a big.LITTLE configuration, with five \textit{big} cores and three \textit{little} cores. The name of each one and their maximum frequency is listed in table \ref{cpuconf}. Of those cores, the three \textit{little} ones are ARM Cortex-A510 cores, the first generation of high efficiency cores from the ARMv9 series of processors running at up to 2 GHz. The \textit{big} cores on the other hand are not five of the same type of processor, instead being made up of three different kinds of ARM core, two Cortex-A710, another two Cortex-A715, and finally one Cortex-X3. Of these \textit{big} cores, the X3 stands out as it is the most powerful one of the set reaching up to 3.36 GHz compared to the A7XX cores that run at up to 2.8 GHz.

\begin{table}[H]
\caption{Kryo CPU configuration}
    \centering
    \begin{tabular}{|c|c|c|}
    \hline
         ID & Type & Frequency (GHz) \\
         \hline
         0 & A510 & 2.016 \\
         1 & A510 & 2.016 \\
         2 & A510 & 2.016 \\
         3 & A710 & 2.803 \\
         4 & A710 & 2.803 \\
         5 & A715 & 2.803 \\
         6 & A715 & 2.803 \\
         7 & X3   & 3.360 \\
         \hline
    \end{tabular}
    \label{cpuconf}
\end{table}

In addition to the CPU, the SoC incorporates a dedicated Qualcomm Adreno 740 GPU, supporting single and half floating-point precision execution (GPU16 and GPU32 execution modes), and a dedicated NPU branded by Qualcomm as the Hexagon Processor, capable of executing both integer and half-precision float computations. Publicly available specifications for the GPU architecture are scarce; the only detailed internal information comes from a third-party blogger's analysis\footnote{Adreno 740 info from Kurnal's blog:\url{https://kurnal.xlog.app/SM8550}}. From this analysis, it is deduced that the GPU comprises 12 Shader Processor units running at up to 719 MHz, a configuration also corroborated by internal Android system configuration files.

The SoC also includes a dedicated NPU featuring specialized units for tensor, scalar, and vector operations, all connected to a shared internal memory. This shared memory is crucial for reducing data transfer latency between these execution units, as the primary feature of the Hexagon is its ability to utilize Micro Tile Inferencing. This technique involves partitioning the AI model layers and dispatching each specific operation (tensor, scalar, or vector) to the corresponding dedicated execution unit for parallel and optimized execution.

\subsection{Software environment}
To fully exploit the available heterogeneous hardware resources, developers utilize the LiteRT framework to integrate AI functionalities into their applications. Since LiteRT natively executes \textit{.tflite} models, a mandatory conversion step is required for models originally developed in frameworks like PyTorch. This conversion is performed externally on a host machine before deployment to the target device. The final \textit{tflite} model is obtained via a multi-step pipeline: first, using PyTorch's export utilities to generate the intermediate \textit{.onnx} format, and subsequently employing Katsuya Hyodo's \textit{onnx2tf} package for conversion to \textit{.tflite}. Although LiteRT offers a direct conversion path via its \textit{ai\_edge\_torc}h package, this feature is still in early development and lacks broad model support; therefore, the \textit{ONNX}-intermediate pipeline was selected. Crucially, the \textit{onnx2tf} toolchain allows for the explicit selection of different model quantization schemes, as summarized in Table~\ref{quants}.

\begin{table}[H]
    \centering
    \caption{Quantization reference}
    \begin{tabular}{|c|c|c|c|c|}
    \hline
         \multirow{2}{*}{Name} & \multicolumn{4}{c|}{Datatype} \\
         \cline{2-5}
          & I/O & Operation & Activation & Weight \\
         \hline
          float32 & FP32 & FP32 & FP32 & FP32\\
          float16 & FP32 & FP32 & FP32 & FP16\\
          int\_quant & FP32 & INT8 & INT8 & INT8 \\
          int\_quant\_int16\_act & FP32 & INT8 & INT16 & INT16 \\
          full\_int\_quant & INT8 & INT8 & INT8 & INT8\\
          full\_int\_quant\_int16\_act & INT16 & INT8 & INT16 & INT16 \\
          dynamic\_range\_quant & FP32 & Mixed & FP32 & Mixed\\
          \hline
    \end{tabular}
    \label{quants}
\end{table}
The quantization employed in this study is post-training quantization. Specifically, static quantization was used, where model weights are converted to the target data type (e.g., INT8) offline, resulting in a model where weights are fixed and directly used in layer operations. In contrast, Dynamic Quantization (represented later as \textit{DYN}) uses "dynamic-range operators" \footnote{Dynamic operators in LiteRT: \url{https://ai.google.dev/edge/litert/models/post_training_quantization\#dynamic_range_quantization}} that only quantize the activations to 8-bit integer precision at runtime, immediately before the computation takes place. The weights themselves are typically stored in 8-bit precision but are often cast back to float precision briefly during computation. This scheme incurs a runtime overhead but is faster than FP32 and often offers better accuracy preservation than fully static methods.

Even though many integer widths can be used for operations, weights and activations when saving the model and are actually supported by the \textit{Tensor} class in LiteRT, there are no optimized implementations of operations for 16 bit integers\footnote{LiteRT documentation states lack of optimized INT16 kernels: \url{https://ai.google.dev/edge/litert/models/post_training_integer_quant_16x8\#evaluate_the_models}}.

In order to facilitate the experimentation a small Android application has been developed that runs the models and stores inference results and speeds.
Starting from LiteRT version 2.0 is planned to change to LiteRT Next, which will make previous applications incompatible due to changes in the API. As of version 2.0.2 of LiteRT, the pipeline to run models in the NPU does not work properly\footnote{LiteRT Commit 34682d3
\url{https://github.com/google-ai-edge/LiteRT/issues/3360\#issuecomment-3335190874}}, so for this paper version 1.4 was also needed to be able to use this accelerator. 

To successfully evaluate performance across the entire heterogeneous architecture, a mixed framework approach was needed due to software compatibility constraints. Specifically, CPU and GPU workloads were executed using the LiteRT Next API, which offered superior control over these units, while NPU workloads required the use of the more stable standard LiteRT framework. This methodological split allows for the comprehensive benchmarking of each accelerator's capabilities, as detailed in Table \ref{lrtversion}.

\begin{table}[H]
    \centering
    \caption{LiteRT version requirements}
    \begin{tabular}{|c|c|}
    \hline
        Device & LiteRT version\\
         \hline
         CPU & \multirow{2}{*}{2.0.2}\\
         GPU & \\
         \cline{2-2}
         NPU & 1.4.0 \\
         \hline
    \end{tabular}
    \label{lrtversion}
\end{table}

 \section{Methodology}
 \label{sec:met}

\begin{figure}[H]
    \centering
    \caption{Experimental Workflow and Model Preparation Pipeline}
    \includegraphics[width=\columnwidth]{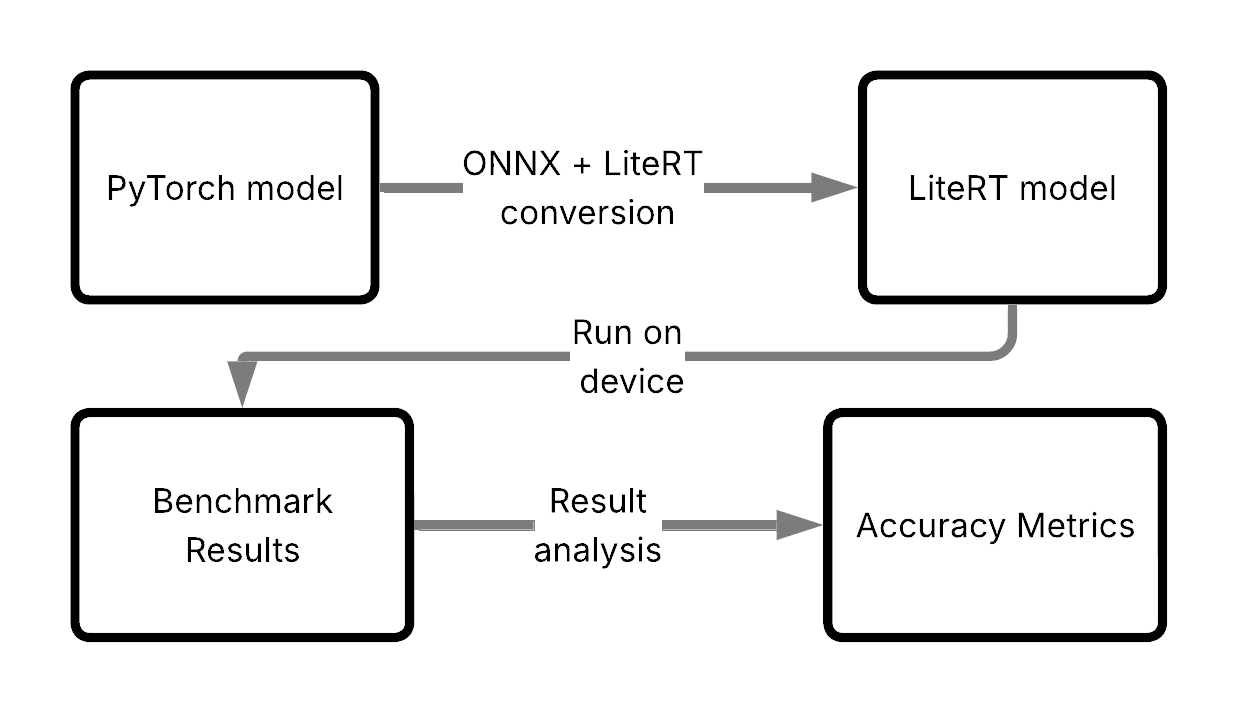}
    \label{fig:workflow}
\end{figure}

 The experimental workflow follows a multi-stage model preparation pipeline, as illustrated in Figure~\ref{fig:workflow} . First, models of all available sizes were sourced from official repositories: the PyTorch \textit{torchvision}\footnote{ResNet models in \textit{torchvision}: \url{https://docs.pytorch.org/vision/main/models/resnet.html}} module for the ResNet family, and the Ultralytics\footnote{Ultralytics' website: \url{https://docs.ultralytics.com/es/models/\#featured-models}} website for the YOLO family. The core preparation pipeline consists of two conversion steps: (1) The PyTorch models are first converted into the intermediate ONNX format using PyTorch's built-in export utilities; (2) the \textit{ONNX} models are then converted to the final TFLite (\textit{.tflite}) format using the \textit{onnx2tf} tool. This last conversion is also where the quantization of the model is performed, by default the \textit{convert} function of \textit{onnx2tf} outputs the model in floating point precision with 32 and 16 bit widths. In order to obtain the other variants shown in table \ref{quants} additional parameters must be passed to the \textit{convert} function specifying that dynamic and integer quantizations are wanted as well\footnote{onnx2tf parameter documentation: \url{https://github.com/PINTO0309/onnx2tf?tab=readme-ov-file\#in-script-usage}}.
 
The benchmarking phase is performed by running inferences against the standard validation datasets corresponding to each model family. Specifically, the ImageNet 1K validation dataset\footnote{ImageNet 1K validation dataset: \url{https://www.kaggle.com/datasets/sautkin/imagenet1kvalid}} is used for ResNet family,while COCO\footnote{COCO validation dataset: \url{http://images.cocodataset.org/zips/val2017.zip}} validation dataset is employed for YOLO family. The primary performance data collected includes the average inference time (latency) across the entire dataset and the corresponding accuracy metric. This comprehensive evaluation was repeated for every model size, quantization variant, and target device.

Finally, the last phase involved the quantitative assessment of the obtained classification and detection results to determine model accuracy. For the YOLO models, the most common accuracy metric is mean average precision (mAP) directly obtained from the Python package \textit{pycocotools}. For ResNet, the top-1 percentage metric is also calculated by computing how many of the total predictions recorded match the ground truth for the image class.

\section{Results}
\label{sec:res}
\input{results}

\section{Conclusions}
\label{sec:conc}
\input{concl}

\balance

\section*{Funding}
This paper has been partially funded by 
the EU (FEDER), the Spanish MINECO under grants PID2021-126576NB-I00 and PID2024-158311NB-I00 funded by MCIN/AEI/10.13039/501100011033 and by European Union ``ERDF A way of making Europe'' and the NextGenerationEU/PRT. 

\bibliographystyle{IEEEtran}
\bibliography{biblio}
\end{document}

%% file: intro.tex
In recent years, Neural Network models have revolutionized numerous domains by providing intelligent solutions for process automation and resource optimization. However, as Artificial Intelligence (AI) systems transition from research prototypes to large-scale commercial applications, computational cost and energy consumption have become major concerns. The inference phase, where models operate continuously, dominates this consumption. Specifically, Google estimates that AI-related tasks accounted for 10-15\% of its total energy use between 2019 and 2021, with inference consuming 60\% of that energy \cite{Zeyu2024}. Similarly, Meta reports an energy capacity distribution of 10:20:70 for experimentation, training, and inference, respectively \cite{wu2022sustainable}. This consumption trend is steep: Google's energy usage grew 21\% annually over the past decade, and Meta's grew even faster at 32\% annually \cite{google2023, meta2023}. 

The escalating demands of AI workloads have driven the evolution of computing personal systems away from homogeneous processors toward heterogeneous architectures. This approach integrates multiple processing units, such as CPUs, GPUs, TPUs, and specialized accelerators, within a single system to optimize performance and energy efficiency by assigning tasks to the unit best suited for them. Furthermore, these advances are critical to the emerging Edge-to-Cloud paradigm \cite{MORESCHINI2024,gill2024modern,AhmedMurtuza2024,fei2023systematic}. In this model, data is processed locally at the "Edge" when possible, optimizing latency and bandwidth, while complex processing is offloaded to the "Cloud".

In the contemporary mobile computing landscape, AI performance is a core differentiator. The increasing complexity of AI models, pushed by the rapid popularization of AI \cite{chatgpt}, demands execution that meets strict real-time and accuracy requirements. Initial solutions involved offloading computation to the GPU, but recently approaches has been move to the use of specific domain-specific-architecture (DSA) as NPUs for this specific task. This DSA is engineered specifically to accelerate the tensor operations that dominate AI model layers, achieving better results in both compute speed and energy efficiency. Along with advancements on hardware, numerous frameworks emerged to allow developers to make use of artificial intelligence in their applications such as PyTorch, ONNX and TensorFlow. However, when mobile devices became powerful enough TensorFlow Lite, an extension of TensorFlow with select operations, enabled local execution on the device itself of certain lightweight models. As of 2024, TensorFlow Lite has been rebranded into LiteRT, short for \textit{Lite} \textit{R}un\textit{T}ime, and in the near future it will be renewed again to LiteRT Next as Google keeps adding functionality to the framework, with the main focus being native support for accelerator offloading of model operations.

In the context of benchmarking, MLPerf~\cite{MLPerf} was created to evaluate Machine Learning (ML) inference systems - from embedded devices to data centers - to decouple software frameworks and architecture neutral. Our work specifically adopts MLPerf principles to evaluate performance on mobile devices based on SoC of a commercial Android device. We selected two industry-standard models that are core components of the MLPerf benchmark suite and are essential for real-world vision systems: ResNet~\cite{ResNet} for Image Classification and the YOLO family~\cite{YOLOreview} for Object Detection. The choice of ResNet and YOLO provides a representative evaluation, contrasting a fundamental classification task with the real-time, complex demands of object detection, thereby generating robust data that reflects the needs of current, energy-efficient applications.

The manuscript is structured as follows: Section~\ref{sec:env} describes topics related to Android's table enviroment used in this work; Section~\ref{sec:met} describes the methodology for this study; Section~\ref{sec:res} presents the results obtained, including performance comparisons; and Section~\ref{sec:conc} summarizes our contributions and proposes future research directions.

%% file: results.tex
Before presenting the results obtained for the ResNet image classification and YOLO object detection models, some observations are necessary. Firstly, the Table~\ref{acro} summarizes the acronyms used throughout the following subsections to refer to variants of precision, device and quantization used in the experimentation. Secondly, the fp32-bit model executed on the CPU on a single thread is considered as a baseline which will serve as reference for speed-up and accuracy loss comparisons. And finally, the accuracy metric depends on the model evaluated: Top 1 accuracy for ResNet, and the \textit{m}ean \textit{A}verage \textit{P}recision (mAP) for YOLO.

\begin{table}[H]
\caption{Acronym correspondence}
    \centering
    \begin{tabular}{|c|c|c|}
    \hline
         &Acronym & Meaning \\
         \hline
         \multirow{6}{*}{Device} & CPU-SC & Single threaded execution \\
         &CPU-MC & Multithreaded execution \\
         &GPU32 & GPU execution with FP32 bit precision\\
         &GPU16 & GPU execution with FP16 bit precision\\
         &NPU & NPU Execution \\
         \hline
         \multirow{7}{*}{Quantization} & FP32 & Floating Point 32 bits\\
         & FP16 & Floating Point 16 bits \\
         & INT8 & Integer 8 bits\\
         & FINT8 & Full Integer 8 bits\\
         & INT16 & Integer 16 bits\\
         & FINT16 & Full Integer 16 bits\\
         & DYN & Dynamic \\
         \hline
    \end{tabular}
    \label{acro}
\end{table}

\subsection{ResNet}
Table \ref{resnet18} presents the inference time measured in milliseconds (ms) for the different quantization and device variants. As the data shows, the 16-bit integer variant exhibits very poor performance (high latency), a trend observed even across different model sizes. Therefore, these quantizations are excluded from the analysis for the remainder of this section. Results also show the clear dominance of the NPU as the fastest executing device, followed by the GPU and lastly the CPU.

\begin{table}[H]
\caption{ResNet18 Average inference time (ms) }
	\centering
	\begin{tabular}{|c|c|c|c|c|c|c|}
    \hline
     & CPU-SC & CPU-MC & GPU32 & GPU16 & NPU\\
    \hline
    FP32   & 79.06 & 26.34 & 13.68 & 5.54 & 1.20 \\
    FP16   & 79.31 & 26.52 & 15.71 & 5.61 & 1.19\\
    INT8   & 23.26 & 5.63 & 21.77 & 22.68 & 0.61\\
\rowcolor{gray!20}
    INT16  & 791.35 & 798.39 & 831.95 & 857.21 & 882.30\\
    FINT8  & 23.06 & 60.36 & 32.57 & 32.48 & 0.51\\
\rowcolor{gray!20}
    FINT16 & 883.34 & 883.18 & 883.50 & 883.16 & 880.98\\
    DYN    & 32.31 & 7.43 & 40.42 &  37.14 & 37.14 \\
    \hline
	\end{tabular}
    \label{resnet18}
\end{table}

Table \ref{resnet18_acc} shows the Top-1 classification accuracy observed during ResNet18 model inference, varying the quantization scheme and the target device. Almost all variants achieve satisfactory accuracy, close to 69\%. The single notable exception is the FINT8 variant, which reaches a drastically low Top-1 accuracy of only 0.08\%. Therefore, this variant is also excluded from the remaining experimentation.

\begin{table}[htb]
	\caption{ResNet18 Top1 accuracy}
	\centering
	\begin{tabular}{|c|c|c|c|c|c|c|}
    \hline
    & CPU-SC & CPU-MC & GPU32 & GPU16 & NPU\\
    \hline
    FP32   & 68.81 & 68.81 & 68.81 & 68.8 & 68.82 \\
    FP16   & 68.82 & 68.82 & 68.82 & 68.79 & 68.81 \\
    INT8   & 65.87 & 65.87 & 65.87 & 65.87 & 65.96 \\
    INT16  & 65.95 & 65.95 & 65.95 & 65.95 & 65.95  \\
   \rowcolor{gray!20}
    FINT8  & 0.08 & 0.08 & 0.09 & 0.09 &  0.08 \\
    FINT16 & 65.95 & 65.95 & 65.95 & 65.95 & 65.95 \\
    DYN    & 68.71 & 68.71 & 68.71 & 68.71 & 68.71 \\
    \hline
	\end{tabular}
    \label{resnet18_acc}
\end{table}

\begin{figure}[H]
\caption{ResNet50 inference speeds across devices \& quantizations}
\centering
\begin{tikzpicture}
\label{resnet50speed}
\begin{axis}[
    ybar,
    bar width=5pt,
    width=\columnwidth,
    height=0.9\columnwidth,
    ylabel={Inference Speed (ms)},
    symbolic x coords={CPU-SC, CPU-MC, GPU32, GPU16, NPU},
    xtick=data,
    legend style={at={(0.5,1.1)}, anchor=north, legend columns=-1},
    ymin=0,
    ytick pos=left, 
    xtick pos = bottom
]
\addplot coordinates {(CPU-SC,126.4) (CPU-MC,61.15) (GPU32, 14.9) (GPU16, 8.1) (NPU, 2.33)};
\addplot coordinates {(CPU-SC,141.05) (CPU-MC,60.91) (GPU32, 14.86) (GPU16, 8.12) (NPU, 2.36)};
\addplot coordinates {(CPU-SC,45.29) (CPU-MC,13.35) (GPU32, 15.17) (GPU16, 8.39) (NPU, 1.04)};
\addplot coordinates {(CPU-SC,47.2) (CPU-MC,17.44) (GPU32, 14.78) (GPU16, 0) (NPU, 53.96)};
\legend{FP32, FP16, INT8, DYN}
\end{axis}
\end{tikzpicture}
\end{figure}
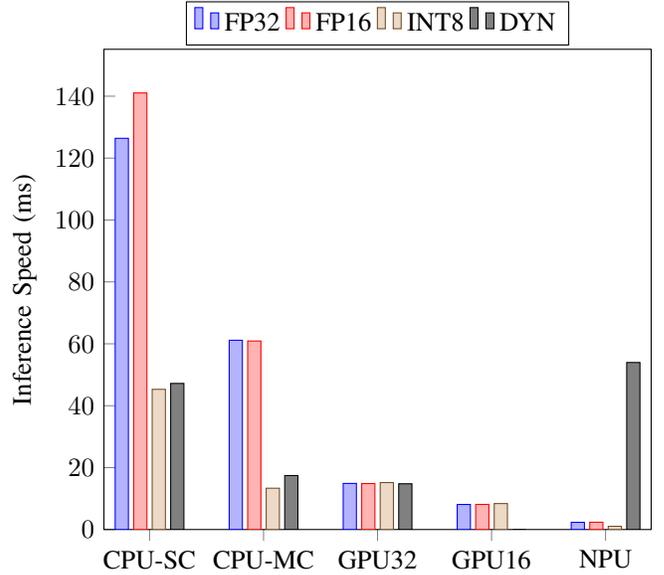

Consistent with the previous results, Figure~\ref{resnet50speed} shows the inference times for the ResNet50 model. Once again, the NPU demonstrates superior speed across the board (INT8 quantization), achieving over $120\times$ faster inference than the CPU-SC. The exception to this performance trend is Dynamic Quantization (DYN), which is significantly slower on the NPU than the regular quantizations. This anomaly is explained by the nature of the dynamically quantized model: it is not uniformly quantized but contains layers of different precisions. While the CPU efficiently handles these mixed precisions, the NPU accelerator lacks native support for such dynamic data type conversion. Consequently, the model must rely on the CPU to perform casting operations, requiring costly data transfers between the NPU and CPU for each precision change. This overhead drastically increases the overall latency, negating the potential speed-up of the lower-precision operations on the accelerator. This outcome reinforces the conclusion that the NPU is optimized for acceleration using data pre-calculated ahead of time, not for operations computed at runtime.

Focusing on CPU performance (see Table~\ref{resnet50mcspeedup}), the analysis revealed that the multithreaded speed-up (CPU-MC, utilizing the 8 available cores) was significantly lower than expected. The highest acceleration was achieved with INT8 quantization, reaching a factor of $3.4\times$, which suggests that the asymmetric multicore architecture significantly limits scalability.

\begin{table}[htb]
\caption{Multithreaded vs single threaded execution (ms) and speed-up in ResNet50}
    \centering
    \begin{tabular}{|c|c|c|c|}
    \hline
         & CPU-SC & CPU-MC & Speed-up \\
         \hline
     FP32 & 126.4 & 61.2 & 2.1x \\
     FP16 & 141.1 & 60.9 & 2.3x \\
     INT8 & 45.3 & 13.4 &  3.4x \\
     DYN  & 47.2 & 17.4 &  2.7x \\
     \hline
    \end{tabular}
    \label{resnet50mcspeedup}
\end{table}

On the GPU side, we observe that quantization hardly affects inference times (see Table \ref{resnet50gpu}). Instead, the largest performance impact stems from the GPU's operating mode (FP32 vs. FP16), resulting in speed-ups approaching $2\times$ if the GPU is configure to operate as GPU16.

\begin{table}[htb]
\caption{GPU FP32 vs FP16 execution (ms) in ResNet50}
    \centering
    \begin{tabular}{|c|c|c|}
    \hline
         & GPU32 & GPU16  \\
         \hline
     FP32 & 14.9 & 8.1  \\
     FP16 & 14.86 & 8.12  \\
     INT8 & 15.17 & 8.39  \\
     DYN  & 14.78 &  8.13 \\
     \hline
    \end{tabular}
    \label{resnet50gpu}
\end{table}

The table \ref{resnet50map} presents accuracy variation when quantizing the model. Using half precision floating point instead of single precision has practically no impact on the accuracy figures. Meanwhile, quantizing the model from fully fp32 to integer or dynamic has a greater impact.
 
\begin{table}[H]
	\caption{ResNet50 Top1 accuracy (\%)}
	\centering
	\begin{tabular}{|c|c|c|c|c|c|c|}
    \hline
    & CPU-SC & CPU-MC & GPU32 & GPU16 & NPU\\
    \hline
    FP32   & 79.09 & 79.09 & 79.06 & 79.09 & 79.1 \\
    FP16   & 79.1 & 79.1 & 79.1 & 79.06 & 79.1 \\
    INT8   & 78.68 & 78.73 & 78.68 & 78.68 & 78.77 \\
    DYN    & 78.9 & 78.91 & 78.9 & - & 78.94 \\
    \hline
	\end{tabular}
    \label{resnet50map}
\end{table}

The analysis of quantization impact reveals a clear correlation between model size and accuracy preservation. As shown in Table~\ref{tab:quantization_loss}, the accuracy loss incurred by INT8 quantization decreases significantly as the model complexity increases (from 2.94\% for ResNet18 down to $\leq 0.5\%$ for larger variants). This trend confirms that larger models mitigate the information loss associated with reduced data width due to their deeper feature extraction capabilities. In contrast, Dynamic Quantization (DYN) consistently introduces negligible loss.

\begin{table}[h]
    \centering
    \caption{Summary of Top-1 Accuracy Loss by Quantization vs. FP32 (\%)}
    \resizebox{\columnwidth}{!}{
    \label{tab:quantization_loss}
    \begin{tabular}{|l|| c c c c c|}
    \hline
        Quant.& ResNet18 & ResNet34 & ResNet50 & ResNet101 & ResNet152 \\ \hline
        INT8  & 2.94\%   & 0.50\%   & 0.41\%   & 0.01\%    & 0.20\% \\
        DYN   & 0.10\%   & 0.04\%   & 0.19\%   & 0.00\%    & 0.07\% \\ 
    \hline
    \end{tabular}
    }
\end{table}

Furthermore, Table~\ref{tab:conversion_loss} demonstrates that converting the original single-precision floating-point model (PyTorch) to the LiteRT framework inherently leads to a conversion-related accuracy loss, regardless of the subsequent quantization applied. This loss ranges between 0.83\% and 1.77\%, suggesting that discrepancies in the inner workings and operation implementations across the frameworks prevent a perfect 1:1 conversion. This factor must be considered when evaluating the overall accuracy degradation.

\begin{table}[h]
    \centering
    \caption{Precision lost (Top-1) due to framework conversion(LiteRT vs. PyTorch) (\%)}
    \resizebox{\columnwidth}{!}{
    \label{tab:conversion_loss}
    \begin{tabular}{|l || c c c c c |}
    \hline
        Model & ResNet18 & ResNet34 & ResNet50 & ResNet101 & ResNet152 \\ \hline
        Lost  & 0.95\%   & 0.83\%   & 1.77\%   & 1.16\%    & 0.98\%    \\ 
    \hline
    \end{tabular}
    }
\end{table}

\subsection{YOLOv8}
The results for YOLOv8 show the same pattern as ResNet: using the accelerators makes the processing much faster. However, YOLOv8 takes longer (has higher latency) than ResNet. This is because YOLOv8 is a more complex network designed for a harder task (object detection) compared to ResNet's simpler one of just image classification.

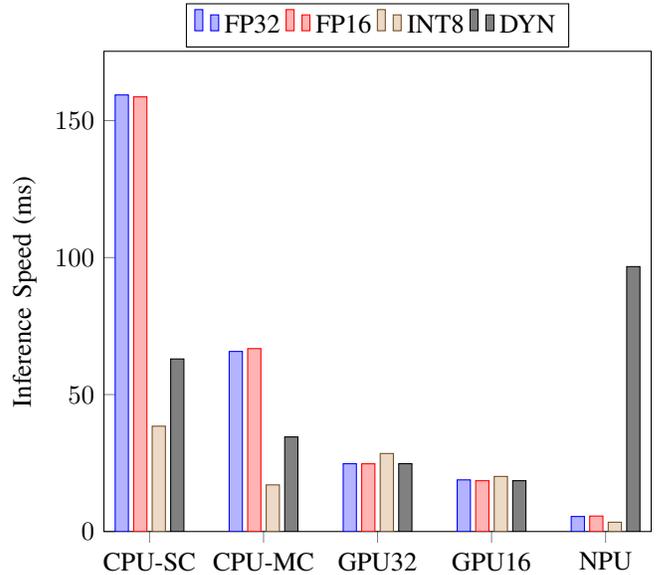
\begin{figure}[H]
\caption{YOLOv8n inference speeds across devices \& quantizations}
\centering
\begin{tikzpicture}
\label{yolov8nspeed}
\begin{axis}[
    ybar,
    bar width=5pt,
    width=\columnwidth,
    height=0.9\columnwidth,
    ylabel={Inference Speed (ms)},
    symbolic x coords={CPU-SC, CPU-MC, GPU32, GPU16, NPU},
    xtick=data,
    legend style={at={(0.5,1.1)}, anchor=north, legend columns=-1},
    ymin=0,
    ytick pos=left, 
    xtick pos = bottom
]
\addplot coordinates {(CPU-SC,159.38) (CPU-MC,65.74) (GPU32, 24.75) (GPU16, 18.86) (NPU, 5.49)};
\addplot coordinates {(CPU-SC,158.68) (CPU-MC,66.78) (GPU32, 24.73) (GPU16, 18.55) (NPU, 5.65)};
\addplot coordinates {(CPU-SC,38.45) (CPU-MC,17.03) (GPU32, 28.48) (GPU16, 20.12) (NPU, 3.40)};
\addplot coordinates {(CPU-SC,63) (CPU-MC,34.54) (GPU32, 24.77) (GPU16, 18.56) (NPU, 96.71)};
\legend{FP32, FP16, INT8, DYN}
\end{axis}
\end{tikzpicture}
\end{figure}

Figure~\ref{yolov8nspeed} shows the inference time involve in YOLO8n model varying devices and quantization. 
The NPU offers the best performance in comparison with devices such as CPU or GPU.

Shifting the focus to accuracy, as can be seen in Table~\ref{tab:yolov8acc}, the integer quantized variants suffer more degradation than in ResNet model. Lower precision, like INT8 quantization, significantly hurts object detection models (YOLOv8) because this task demands more information (localization and classification) than simpler tasks like image classification. The dynamic quantization (DYN) method performs much better, losing a maximum of only 0.1 mAP across the models, which is negligible. Conversely, INT8 quantization causes a large performance drop, averaging about 6.5 mAP points lost across all model sizes.

\begin{table}[h]
    \centering
    \caption{YOLOv8 accuracy loss (mAP) compared to FLOAT32 precision}
    \resizebox{\columnwidth}{!}{
    \label{tab:yolov8acc}

    \begin{tabular}{|l || c c c c c |}
    \hline
Quant. & YOLOv8n & YOLOv8s & YOLOv8m & YOLOv8l & YOLOv8x \\ \hline
INT8   & 6.5     & 6.2     & 6.2     & 7.5    & 6.1\\
DYN    & 0.1     & 0.0     & 0.0     & 0.1    & 0.1\\
    \hline
    \end{tabular}
   }
\end{table}

\subsection{YOLO11}

\begin{figure}[H]
\caption{YOLO11n inference speeds across devices \& quantizations}
\centering
\begin{tikzpicture}
\label{yolo11nspeed}
\begin{axis}[
    ybar,
    bar width=5pt,
    width=\columnwidth,
    height=0.9\columnwidth,
    ylabel={Inference Speed (ms)},
    symbolic x coords={CPU-SC, CPU-MC, GPU32, GPU16, NPU},
    xtick=data,
    legend style={at={(0.5,1.1)}, anchor=north, legend columns=-1},
    ymin=0,
    ytick pos=left, 
    xtick pos = bottom
]
\addplot coordinates {(CPU-SC,124.3) (CPU-MC,57.02) (GPU32, 24.03) (GPU16, 18.20) (NPU, 6.03)};
\addplot coordinates {(CPU-SC,124.97) (CPU-MC,57.25) (GPU32, 23.98) (GPU16, 18.24) (NPU, 5.99)};
\addplot coordinates {(CPU-SC,33.98) (CPU-MC,17.3) (GPU32, 27.56) (GPU16, 20.72) (NPU, 3.84)};
\addplot coordinates {(CPU-SC,62.82) (CPU-MC,39.67) (GPU32, 23.87) (GPU16, 18.21) (NPU, 0)};
\legend{FP32, FP16, INT8, DYN}
\end{axis}
\end{tikzpicture}
\end{figure}
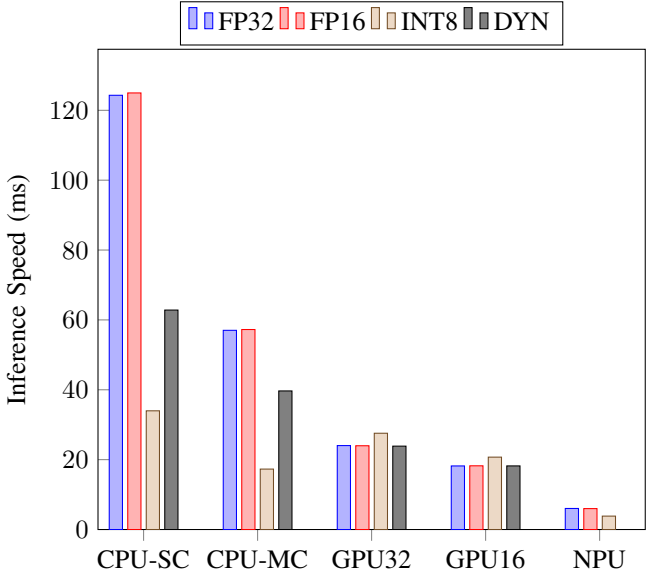

With YOLO11 being an upgraded version of YOLOv8 it is expected to behave similarly but come with some improvements compared to its predecessor. The Figure~\ref{yolo11nspeed} presents the newer YOLO11 inference times. It generally behaves like YOLOv8 but shows minor changes: CPU inference is slightly faster, while GPU speeds are nearly identical. Execution on the NPU is marginally slower for both float and INT8 models, and dynamic quantization fails to run entirely on the NPU. This performance shift is likely due to YOLO11's more complex architecture and new layers compared to its predecessor.

Accuracy results for YOLO11 match the general trend seen in YOLOv8 as shown in Table~\ref{tab:yolo11acc}. However, integer quantization (INT8) in YOLO11 models causes a slightly greater loss in accuracy, averaging around 7.2 mAP points across the models. As with the previous version, dynamic quantization (DYN) maintains an almost perfect accuracy result, with a maximum loss of just 0.2 mAP.

\begin{table}[h]
    \centering
    \caption{YOLO11 accuracy loss (mAP) compared to FLOAT32 precision}
    \resizebox{\columnwidth}{!}{
    \label{tab:yolo11acc}
    \begin{tabular}{|l || c c c c c |}
    \hline
    Quant. & YOLO11n & YOLO11s & YOLO11m & YOLO11l & YOLO11x \\ \hline
    INT8 & 6.6 & 6.8 & 6.7 & 8.7 & 7.2 \\
    DYN & 0.1 & 0.1 & 0.1 & 0.2 & 0.1  \\
    \hline
    \end{tabular}
    }
\end{table}

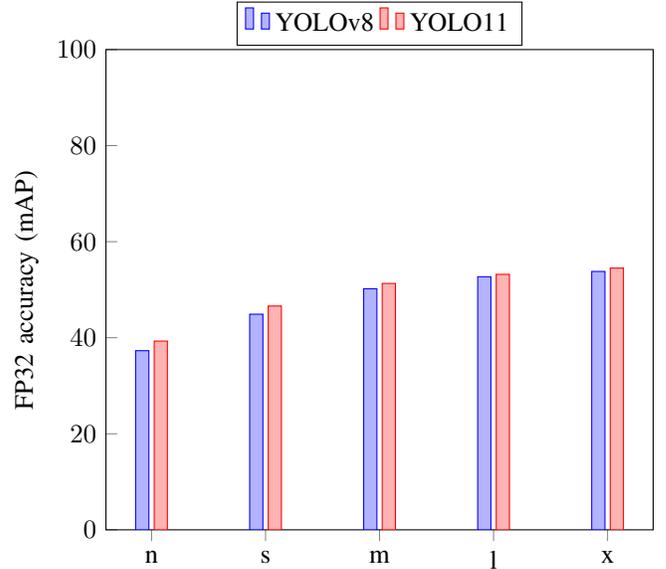
\begin{figure}[H]
\caption{YOLO11 vs YOLOv8 accuracy comparison}
\centering
\begin{tikzpicture}
\label{yolo11v8}
\begin{axis}[
    ybar,
    bar width=5pt,
    width=\columnwidth,
    height=0.9\columnwidth,
    ylabel={FP32 accuracy (mAP)},
    symbolic x coords={n, s, m, l, x},
    xtick=data,
    legend style={at={(0.5,1.1)}, anchor=north, legend columns=-1},
     xticklabel style={
    rotate=0,
    anchor=north,
},
    ymin=0,
    ymax=100,
    ytick pos=left, 
    xtick pos = bottom
]
\addplot coordinates {(n,37.3) (s,44.9) (m, 50.2) (l, 52.7) (x, 53.8)};
\addplot coordinates {(n,39.3) (s,46.6) (m, 51.3) (l, 53.2) (x, 54.5)};
\legend{YOLOv8,YOLO11}
\end{axis}
\end{tikzpicture}
\end{figure}

Finally, it isimportant to note that YOLO11 exhibits a minor degradation in accuracy compared directly against its original PyTorch version, caused by the conversion to the required format for the LiteRT framework. This nominal loss, typically ranging between 0.2 and 0.4 mAP percentage points, implies that the novel layers incorporated into the YOLO11 architecture present a greater challenge for effective conversion into the LiteRT runtime environment. Despite this minor conversion-induced discrepancy, the empirical evidence confirms that YOLO11 provides a superior overall performance in FP32 accuracy across all model scales when benchmarked against YOLOv8 (as validated by the data presented in Figure \ref{yolo11v8}). Thus, YOLO11 establishes itself as the more precise model for object detection tasks.

\subsection{Discussion}
After analyzing the results obtained from all the executions of the different models in the various devices, the most suitable accelerator is the NPU, which achieves impressive speed-ups in the smaller models
approximately $47\times$ in YOLOv8n and $130\times$ in ResNet18, and this ratio generally increases with the model size, reaching up to $298\times$ speed-up in YOLOv8x versions. Table~\ref{NPUspeedup} summarizes the speed-up achieved for each model and quantization variant.

\begin{table}[H]
\caption{NPU execution speed-ups}
    \centering
    \begin{tabular}{|c|c|c|c|c|c|}
    \hline
         \multicolumn{2}{|c|}{} & FP32 & FP16 & INT8 & DYN \\
         \hline
\multirow{5}{*}{Resnet}& 18 & 65.9x & 66.4x & 129.6x & 2.1x \\
                       & 34 & 89.2x & 88.2x & 190x & 3.7x \\
                       & 50 & 54.3x & 53.3x & 121.5x & 2.3x \\
                       & 101 & 64.9x & 63.6x & 149.7x & 3x \\
                       & 152 & 70x & 67.5x & 164.1x & - \\
    \hline
\multirow{5}{*}{YOLOv8}& n & 29x & 28.2x & 46.8x & 1.6x \\
                       & s & 60.2x & 60.5x & 128.3x & 2.9x \\
                       & m & 77.6x & 81.6x & 209.4x & - \\
                       & l & 99.5x & 100.7x & 277.6x & - \\
                       & x & 98.2x & 100x & 298x & - \\
    \hline
\multirow{5}{*}{YOLO11}& n & 20.6x & 20.8x & 32.4x & - \\
                       & s & 44.2x & 41.5x & 76.5x & - \\
                       & m & 62.3x & 64.9x & 146.1x & - \\
                       & l & 71.2x & 72.5x & 160.3x & - \\
                       & x & 72.9x & 72.6x & 173.9x & - \\
    \hline
    \end{tabular}
    \label{NPUspeedup}
\end{table}

Another important observation is that the dynamically quantized model seems to have some troubles across all models, due to reduced speed-ups in comparison with other quantization options or even failure to run. Scalability is also observed when model size increases as acceleration grows, being more impressive for YOLO8 family.

Focusing on CPU performance, Table~\ref{mtspeedup} clearly demonstrates that the maximum achievable speed-up with multi-threading is predominantly driven by integer precision (INT8), consistently ranging between $10\times$ and $15\times$ faster execution, while floating-point precisions achieves a modest $2\times$ to $3\times$ acceleration. The multi-threading efficiency varies by model family and size: ResNet models exhibit a decreased speed-up in INT8 as model size increases, whereas YOLO models demonstrate the opposite trend, achieving a better speed-up as they scale up.

\begin{table}[H]
    \caption{Multithreaded execution speed-ups}
    \centering
    \begin{tabular}{|c|c|c|c|c|c|}
    \hline
         \multicolumn{2}{|c|}{} & FP32 & FP16 & INT8 & DYN \\
         \hline
\multirow{5}{*}{Resnet}& 18 & 3x & 3x & 14x & 10.6x \\
                       & 34 & 3.1x & 3.1x & 15.1x & 12x \\
                       & 50 & 2x & 2x & 9.5x & 7.2x \\
                       & 101 & 2.2x & 2.1x & 9.2x & 7.2x \\
                       & 152 & 2.2x & 2.3x & 9.4x & 7.6x \\
    \hline
\multirow{5}{*}{YOLOv8}& n & 2.4x & 2.4x & 9.4x & 4.6x \\
                       & s & 2.3x & 2.3x & 11.1x & 6.7x \\
                       & m & 2.4x & 2.4x & 12.5x & 7.5x \\
                       & l & 2.3x & 2.7x & 12.8x & 9.7x \\
                       & x & 2.7x & 2.1x & 13.4x & 10.1x \\
    \hline
\multirow{5}{*}{YOLO11}& n & 2.2x & 2.2x & 7.2x & 3.1x \\
                       & s & 2.1x & 2.1x & 10.4x & 4.3x \\
                       & m & 2.4x & 2.3x & 13.1x & 6.2x \\
                       & l & 2.6x & 2.3x & 13.4x & 6.6x \\
                       & x & 2.2x & 2.1x & 12.5x & 7.3x \\
    \hline
    \end{tabular}
    
    \label{mtspeedup}
\end{table}

Finally, the analysis on GPU device presented in the Table~\ref{gpupeedup} reveals that the best performance rates are found in GPU16 mode. In contrast to the CPU, variations in precision (FP32, FP16, INT8, DYN) have a marginal impact on the speed-up factor, suggesting that performance is constrained other factors such as memory bandwidth or arithmetic complexity. Nevertheless, results suggests that FP16 is frequently the optimal choice, effectively balancing speed and precision on the GPU. Regarding scaling, the YOLO models demonstrate superior efficiency on the GPU, with the largest models (reaching up-to $39\times$ for YOLOv8x) achieving the highest acceleration, while ResNet models maintain a more consistent speed-up across their sizes. Speedup results indicates that larger models take more advantage of being executed at half precision than smaller models, this is in accordance with how GPUs behave in regular tasks, as to take full advantage of this device the workload needs to be massively parallel like with these larger models.

\begin{table}[H]
    \caption{GPU16 execution speed-ups}
    \centering
    \begin{tabular}{|c|c|c|c|c|c|}
    \hline
         \multicolumn{2}{|c|}{} & FP32 & FP16 & INT8 & DYN \\
         \hline
\multirow{5}{*}{Resnet}& 18 & 17.5x & - & 16.9x & - \\
                       & 34 & 20x & 20x & - & 20x \\
                       & 50 & 15.6x & 15.6x & 15.1x & 15.6x \\
                       & 101 & 15.8x & 15.8x & 15.5x & 15.8x \\
                       & 152 & 16.5x & 16.5x & 16.3x & 16.3x \\
    \hline
\multirow{5}{*}{YOLOv8}& n & 8.5x & 8.6x & 8x & 8.5x \\
                       & s &  17.2x &17.1x & 15.5x & 17.1x \\
                       & m & 14.6x & 14.6x & 13.1x & 14.6x \\
                       & l & 35x & 34.8x & 31.9x & 34.9x \\
                       & x & 38.3x & 38.2x & 34.8x & 38.6x \\
    \hline
\multirow{5}{*}{YOLO11}& n & 6.8x & 6.8x & 6x & 6.8x \\
                       & s & 14x & 14x & 12.3x & 14x \\
                       & m & 23.2x & 23.2x & 19.3x & 23.1x \\
                       & l & 25.5x & 25.2x & 22x & 25.2x \\
                       & x & 30.1x & 30.1x & 26.5x & 30.2x \\
    \hline
    \end{tabular}
    
    \label{gpupeedup}
\end{table}

Finally, selecting the optimal configuration is inherently a challenge due to the conflicting objectives typically sought, such as achieving high accuracy alongside moderate inference times without ignoring limited power consumption constraints. To address this multi-criteria decision problem, a multi-objective optimization approach using a Pareto Front analysis is proposed. In order to avoid cluttering, only FP16 and INT8 quantizations were selected to be shown in the graphs as they are involved in the most promising configurations.

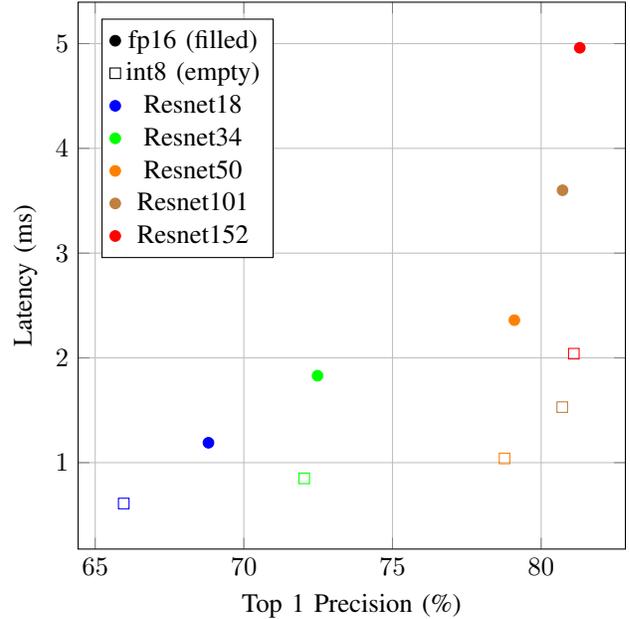
\begin{figure}[H]
\caption{Pareto Front ResNet}
\begin{tikzpicture}
\label{fig:ParetoResnet}
\begin{axis}[
    xlabel={Top 1 Precision (\%)},
    ylabel={Latency (ms)},
    grid=both,
    width=\columnwidth,
    height=\columnwidth,
legend style={at={(0.2,0.97)}, anchor=north},
]
\addlegendimage{only marks, mark=*, black}
\addlegendentry{fp16 (filled)}
\addlegendimage{only marks, mark=square, black}
\addlegendentry{int8 (empty)}
\addplot[only marks,mark=*,blue] coordinates {(68.81,1.19)};
\addlegendentry{Resnet18}
\addplot[only marks,mark=*,green] coordinates {(72.48,1.83)};
\addlegendentry{Resnet34}
\addplot[only marks,mark=*,orange] coordinates {(79.1,2.36)};
\addlegendentry{Resnet50}
\addplot[only marks,mark=*,brown] coordinates {(80.72,3.6)};
\addlegendentry{Resnet101}
\addplot[only marks,mark=*,red] coordinates {(81.3,4.96)};
\addlegendentry{Resnet152}
\addplot[only marks,mark=square,blue, fill=white] coordinates {(65.96,0.61)};
\addplot[only marks,mark=square,green] coordinates {(72.03,0.85)};
\addplot[only marks,mark=square,orange] coordinates {(78.77,1.04)};
\addplot[only marks,mark=square,brown] coordinates {(80.71,1.53)};
\addplot[only marks,mark=square,red] coordinates {(81.1,2.04)};
\end{axis}
\end{tikzpicture}
\end{figure}

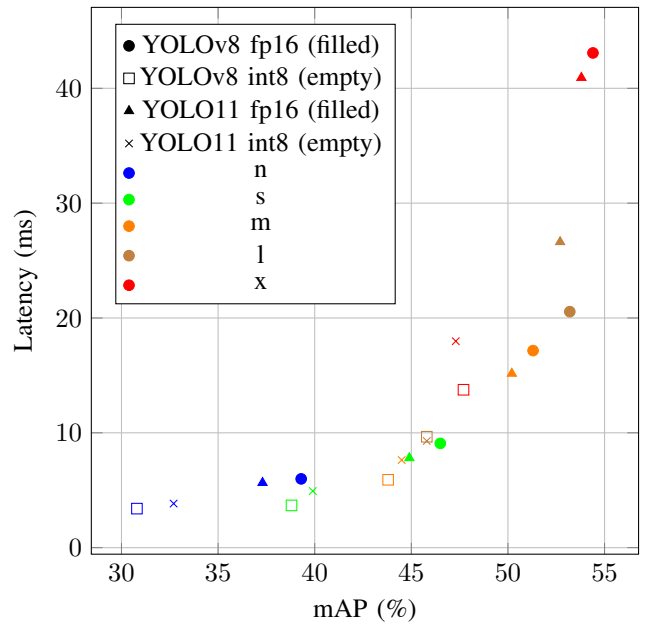
\begin{figure}[H]
    \centering
    \caption{Pareto Front YOLO}
    \begin{tikzpicture}
\label{fig:ParetoYOLO}
\begin{axis}[
    xlabel={mAP (\%)},
    ylabel={Latency (ms)},
    title={},
    grid=both,
    width=\columnwidth,
    height=\columnwidth,
  legend style={at={(0.3,0.97)}, anchor=north},
]
\addlegendimage{only marks, mark=*, black}
\addlegendentry{YOLOv8 fp16 (filled)}
\addlegendimage{only marks, mark=square, black}
\addlegendentry{YOLOv8 int8 (empty)}
\addlegendimage{only marks, mark=triangle*, black}
\addlegendentry{YOLO11 fp16 (filled)}
\addlegendimage{only marks, mark=x, black}
\addlegendentry{YOLO11 int8 (empty)}
\addlegendimage{only marks, mark=*, blue}
\addlegendentry{n}
\addlegendimage{only marks, mark=*, green}
\addlegendentry{s}
\addlegendimage{only marks, mark=*, orange}
\addlegendentry{m}
\addlegendimage{only marks, mark=*, brown}
\addlegendentry{l}
\addlegendimage{only marks, mark=*, red}
\addlegendentry{x}
\addplot[only marks,mark=triangle*,blue] coordinates {(37.3,5.65)};
\addplot[only marks,mark=square,blue] coordinates {(30.8,3.4)};
\addplot[only marks,mark=triangle*,green] coordinates {(44.9,7.81)};
\addplot[only marks,mark=square,green] coordinates {(38.8,3.68)};
\addplot[only marks,mark=triangle*,orange] coordinates {(50.2,15.16)};
\addplot[only marks,mark=square,orange] coordinates {(43.8,5.91)};
\addplot[only marks,mark=triangle*,brown] coordinates {(52.7,26.62)};
\addplot[only marks,mark=square,brown] coordinates {(45.8,9.66)};
\addplot[only marks,mark=triangle*,red] coordinates {(53.8,40.92)};
\addplot[only marks,mark=square,red] coordinates {(47.7,13.75)};
\addplot[only marks,mark=*,blue] coordinates {(39.3,5.99)};
\addplot[only marks,mark=x,blue] coordinates {(32.7,3.84)};
\addplot[only marks,mark=*,green] coordinates {(46.5,9.08)};
\addplot[only marks,mark=x,green] coordinates {(39.9,4.93)};
\addplot[only marks,mark=*,orange] coordinates {(51.3,17.16)};
\addplot[only marks,mark=x,orange] coordinates {(44.5,7.63)};
\addplot[only marks,mark=*,brown] coordinates {(53.2,20.55)};
\addplot[only marks,mark=x,brown] coordinates {(45.8,9.29)};
\addplot[only marks,mark=*,red] coordinates {(54.4,43.08)};
\addplot[only marks,mark=x,red] coordinates {(47.3,17.98)};
\end{axis}
\end{tikzpicture}
\end{figure}
Figures~\ref{fig:ParetoResnet} and~\ref{fig:ParetoYOLO} present the resulting Pareto Front, which displays the non-dominated trade-offs between accuracy and inference time across the two selected quantization schemes and deployment devices. This analysis emphasizes the best possible compromises available to system designers by clearly distinguishing between superior and sub-optimal configurations. It can be deduced that in the case of ResNet (see Figure~\ref{fig:ParetoResnet}) INT8 is clearly the way to go, as the latency decrease compensates the accuracy loss of using integer precision. In contrast, for YOLO (Figure~\ref{fig:ParetoYOLO}), though, as these models suffer much more from integer related accuracy loss, these variants of the model are further away from the optimum front. Another key takeaway is that YOLO11 seems to outperform YOLOv8 in the smaller sizes where the accuracy increase makes it fare better at the cost of slightly increased latency, this advantage goes away the bigger the model is and consequently the slower it becomes.

%% file: concl.tex

This paper evaluated the performance of ResNet and YOLO models for image classification and object detection, respectively, across three devices (CPU, GPU, NPU) and four precisions in an Android tablet. The key findings are:
\begin{itemize}
\item The NPU is the best execution device, achieving speed-ups up to $120\times$ compared to the CPU single-core baseline, confirming its essential role in low-latency edge AI.
\item Optimal Quantization is a trade-off, and the choice between FP16 and INT8 depends on the model:
      \begin{itemize}
      \item INT8 is the optimal precision in ResNet, as its speed gain on the NPU over the CPU outweighs its minimal accuracy loss.
      \item Quantization provides negligible speed benefits over FP16 execution in GPU, making it the preferred choice for balancing speed and full accuracy on the GPU.
      \end{itemize}
\item Regarding model Performance \& Scaling:
      \begin{itemize}
      \item YOLO11s represents a strong candidate on the Pareto Front, offering a superior speed/accuracy compromise for object detection with FP16 quantization.
      \item YOLO11 provides an overall accuracy gain over YOLOv8 for smaller model sizes, despite its slightly increased complexity.
      \end{itemize}
\end{itemize}
From the system limitations perspective, we have identified that Dynamic Quantization (DYN) fails on the NPU due to lack of native support, and CPU multi-threading scalability is poor ($\approx 3.4\times$ max speed-up) due to the asymmetric core architecture.
As future work, we plan to complete a Pareto Analysis that integrates power consumption to finalize the multi-objective optimization (accuracy, latency, power efficiency) and to mitigate NPU issues such as DYN compatibility to eliminate the $\approx 1\%$ framework conversion accuracy loss observed with LiteRT. Furthermore, we intend to generalize these findings by studying the behaviour of additional models included in the MLPerf benchmark suite, specifically those designed for tasks such as natural language processing or image segmentation.